\documentclass{article}

\PassOptionsToPackage{numbers, compress}{natbib}
\usepackage[final]{nips_2017}

\usepackage[utf8]{inputenc} 
\usepackage[T1]{fontenc}    
\usepackage{hyperref}       
\usepackage{url}            
\usepackage{booktabs}       
\usepackage{amsfonts}       
\usepackage{nicefrac}       
\usepackage{microtype}      
\usepackage[table]{xcolor}

\newcommand{\talktome}{Talk2Me}
\title{On the importance of normative data in speech-based assessment}

\author{
  Zeinab Noorian\\
  Department of Computer Science\\
  University of Toronto\\
  Toronto, Canada \\
  \texttt{znoorian@cs.toronto.edu} \\
 \And
 Chlo\'{e} Pou-Prom \\
 Department of Computer Science \\
University of Toronto \\
Toronto, Canada \\
 \texttt{chloe@cs.toronto.edu} \\
 \And
 Frank Rudzicz \\
 Toronto Rehabilitation Institute,  University Health Network; \\
 Department of Computer  Science, University of Toronto; and \\
 Vector Institute for Artificial Intelligence\\
 Toronto, Canada\\
 \texttt{frank@cs.toronto.edu} \\
}

\begin{document}

\maketitle

\begin{abstract}
Data sets for identifying Alzheimer's disease (AD) are often relatively sparse, which limits their ability to train generalizable models. Here, we augment such a data set, DementiaBank, with each of two normative data sets, the Wisconsin Longitudinal Study and Talk2Me, each of which employs a speech-based picture-description assessment. Through minority class oversampling with ADASYN, we outperform state-of-the-art results in binary classification of people with and without AD in DementiaBank. This work highlights the effectiveness of combining sparse and difficult-to-acquire patient data with relatively large and easily accessible normative datasets.
\end{abstract}

\section{Introduction}

Alzheimer's Disease (AD) is a neurodegenerative disease which affects 5.5 million Americans and whose care cost \$259 billion in the United States in 2017 \cite{Association2017}. Despite its prevalence, it can be challenging to recruit participants with cognitive decline for research studies, due to issues ranging from ethics protocol restrictions for vulnerable populations to caregiver fatigue. Datasets for AD are therefore often sparse \cite{Rentoumi2014}.

Language decline is one of the main symptoms of AD and several studies have consequently applied natural language processing and machine learning to quantify differences between AD and healthy speech. Wankerl {\em et al} \cite{Wankerl2017} used a simple N-gram based approach to build language models for control participants and AD patients. Using a perplexity measure, they achieved a classification result of 77.1\%.  Rentoumi {\em et al} \cite{Rentoumi2014} considered a slightly more challenging task, using frequency unigrams to differentiate between picture descriptions from AD participants with and without additional vascular pathology ($N=18$ in each group); their highest accuracy was 75\%. Other approaches have considered a greater number of features. Guinn {\em et al} \cite{Guinn2012} distinguished between AD and healthy language samples, with up to 79.5\% accuracy, using 80 conversations and features such as filled pauses, repetitions, and incomplete words. Meilan {\em et al} \cite{Meilan2014} distinguished between 30 patients with AD and 36 healthy controls, obtaining an accuracy of 84.8\%, using temporal and acoustic features such as percentage and number of voice breaks, shimmer, and noise-to-harmonics ratio. Similarly, Fraser {\em et al} detected primary progressive aphasia \cite{Fraser:2013} and AD \cite{Fraser2015} with up to 100\% and 82\% accuracy, respectively, using a wide array of lexicosyntactic and acoustic features during story retelling and picture description texts. 

Previous experiments on classifying between AD and healthy speech highlight the need for bigger datasets. For example, Andrade De Oliveira {\em et al} \cite{AndradeDeOliveira2014} used normative data to identify AD patients, albeit on neuroimaging data. We take a similar approach and explore combining normative data with existing speech and text transcripts collected on participants with AD in a picture description task. By combining synthetic sampling and normative data, we obtain state-of-the-art results on the DementiaBank dataset.

\section{Data}
Here, we combine a dataset containing AD participants, DementiaBank, with each of two normative datasets, consisting of only healthy participants, from the Wisconsin Longitudinal Study and the \talktome\ project. Table \ref{tab:demog} shows demographics for these datasets, each of which employs the same `Cookie Theft' picture description task from the Boston Diagnostic Aphasia Examination \cite{Goodglass1983}.

\subsection{DementiaBank (DB)}
In DB, which is part of the TalkBank project \cite{MacWhinney11}, each participant was above 44 years old, and had at least 7 years of education. Participants also had no history of nervous system disorders, had an initial Mini-Mental State Exam (MMSE) score of 10 or greater\footnote{The MMSE is administered by a clinician and measures cognitive ability. MMSE scores range from 0 to 30, with scores greater than 23 being associated with no cognitive decline. Scores between 18 and 23 are associated with mild cognitive impairment, scores of 11 to 17 correspond to a moderate degree of impairment, and scores of 10 or below are associated with severe cognitive impairment \cite{Folstein1975}.}, and were able to give informed consent \cite{Becker1994}. Each participant was assigned to either the `Dementia' group ($N=167$) or the `Control' group ($N=97$) based on their medical histories and an extensive neuropsychological and physical assessment battery. Additionally, since many subjects repeated their engagement at yearly intervals (up to five years), we use $240$ samples from those in the `Dementia' group, and $233$ from those in the `Control' group. Each speech sample was recorded and manually transcribed at the word level following the CHAT protocol \cite{MacWhinney1992}. Narratives were segmented into utterances and annotated with filled pauses, paraphasias, and unintelligible words. 

\subsection{\talktome \ (T2M)}\label{sec:t2m}

 \talktome\ is an online language assessment from the University of Toronto\footnote{\url{https://www.cs.toronto.edu/talk2me}}. It consists of seven tasks, including picture descriptions, story retellings, word-colour Stroop, fluency tasks, and self-reported evaluation of mood. The tasks are performed online, with participants entering their answers through text or through speech recordings. Answers to the picture description task are collected as audio recordings. Participants are shown a random picture during each session, including pictures from Flickr, the Webber Photo Cards: Story Starters collection \cite{Webber2005}, and Cookie Theft. Crucially, unlike DB, no human-produced transcripts are included. We therefore apply the Kaldi open-source automatic speech recognition (ASR) engine \cite{Povey2011}, using a long short-term memory network with i-Vector input \cite{Verma2015} and a reverberation model, trained on the Fisher data \cite{Cieri2005}. Our {\em ad hoc} evaluation of a random portion of these data suggests a word-error rate of approximately 12.5\%.

\subsection{Wisconsin Longitudinal Study (WLS)}

The second normative dataset is the Wisconsin Longitudinal Study (WLS), which is recorded over several decades on a 1/3 random sample of all Wisconsin high school graduates in 1957 ($N=10,317$) born between 1938 and 1940 \cite{Herd2014}. Survey data were collected from the original respondents or their parents in 1957, 1964, 1975, 1992, 2004, and 2011, and participants performed the `Cookie Theft' picture description task in the 2011 survey. Only the audio was retained from that survey, so we therefore apply the same Kaldi-based ASR engine to these data as we do in Section \ref{sec:t2m}.

\begin{table}[h]
\caption{\label{tab:demog} Demographics for the three data sets for patients with AD and controls (CT). Years are indicated by their means and standard deviations. }
\centering
\begin{tabular}{|r| rr | rr | rr | rr}
\hline
& \multicolumn{2}{|c|}{Sex (M/F)} & \multicolumn{2}{|c|}{Age (years)} & \multicolumn{2}{|c|}{Education (years)}\\
& AD & CT & AD & CT & AD & CT\\
\hline
DB & 82/158 & 82/151 & 71.8 (8.5) & 65.2 (7.8) & 12.5 (2.9) & 14.1 (2.4)\\
\talktome & 0/0 & 187/118 & - (-) & 27.7 (10.1) & - (-) & 16.7 (2.0) \\
WLS & 0/0 & 681/685 & - (-) & 71.2 (4.4) & - (-) & 13.7 (2.0) \\    
\hline
\end{tabular}

\end{table}

\section{Experiments}\label{experiments}

From the picture description transcripts, we extract 567 features, including various lexical features (e.g., mean number of syllables per word,  mean word length, various parts-of-speech and phrase type counts and ratios), and syntactic features (e.g., ratios of various context-free grammatical constructions, and the total number of T-units). We also compute vocabulary similarity with the cosine distance between words. Finally, we compute various subjective measures, including the Flesch-Kincaid score for readability \cite{Kincaid1975}, LIWC psycholinguistic features \cite{Pennebaker2015}, and valence from the Stanford Sentiment Analyzer \cite{Socher2013}.

We then perform a one-way ANOVA and retain the features with $p$-values $\leq$ 0.005, set empirically. In the binary classification task, this selects 142 features with DB only, 311 features with DB + WLS, and 364 features with DB + T2M. In the multi-class classification task, we use 174 features with DB only, 293 features with DB + WLS, and 361 features with DB + T2M. A subset of the top features identified by the ANOVA test for differentiating between CT and AD participants are presented in Table \ref{features}.

\begin{table}[h]
\caption{Top features for  DB, DB + T2M, and DB + WLS, following a one-way ANOVA test for differentiating between AD vs CT participants.}
\label{features}
\centering
\begin{tabular}{|c|c|c|}\hline
\textbf{DB} & \textbf{DB + T2M} & \textbf{DB + WLS} \\ \hline
\cellcolor{red!50}Mean syl/word & \cellcolor{green!75} Mean cosine distance &	 \cellcolor{green!75} Mean cosine distance \\ \hline
\cellcolor{yellow!} Flesch & \cellcolor{blue!25} $ROOT\  \rightarrow \ S$	& \cellcolor{blue!25} $ROOT\  \rightarrow \ S$ \\  \hline
\cellcolor{red!50}Mean word length &	\cellcolor{blue!25}T&	\cellcolor{blue!25}S\\  \hline
\cellcolor{yellow!} Receptiviti $family\_oriented$	 & \cellcolor{blue!25}S	& \cellcolor{green!75}  Min. cosine distance\\  \hline
\cellcolor{blue!25} $NP \ \rightarrow \ PRP$ & \cellcolor{yellow!} Receptiviti food\_focus & \cellcolor{blue!25}$S \rightarrow \ CC\,\,\,NP\,\,\,VP$\\  \hline
\cellcolor{blue!25} $ADVP \ \rightarrow  \ RB$	& \cellcolor{yellow!} LIWC $Apostro$ &	\cellcolor{blue!25} $S \rightarrow ADVP\,\,\,NP\,\,\,VP$\\  \hline
\cellcolor{yellow!} Flesch-Kinkaid	& \cellcolor{green!75}  Min. cosine distance & $ \cellcolor{blue!25}S \ \rightarrow \ NP\,\,\,VP$ \\  \hline
\cellcolor{yellow!} Receptiviti $sociable$ &	\cellcolor{yellow!} LIWC $ingest$ &	\cellcolor{green!75}  Cosine cutoff (0.5) \\  \hline
\cellcolor{red!50}Adverbs & \cellcolor{blue!25} $S \ \rightarrow  \ ADVP\,\,\,NP\,\,\,VP$ & \cellcolor{blue!25} $VP \ \rightarrow \ VBZ\,\,\,VP$ \\  \hline
\cellcolor{red!50}PRP ratio & \cellcolor{yellow!} Mean Stanford sentiment (negative) & \cellcolor{blue!25}T \\  \hline
\end{tabular}

\begin{tabular}{|c|c|c|c|}\hline
\cellcolor{red!50}Lexical& \cellcolor{blue!25} Syntactic & \cellcolor{green!75}Semantic similarity & \cellcolor{yellow!}Subjective\\
\hline
\end{tabular}

\end{table}

We combine DB data with each normative dataset, WLS or T2M, in turn.  To avoid bias introduced by class imbalance, we oversample the minority class with ADASYN \cite{He2008}. ADASYN extends methods such as SMOTE by synthesizing points closer to the decision boundary. Data were randomly split 80/20 for training/testing, ensuring each participant's samples do not occur in both sets. We apply ADASYN on the training set only.

We consider a random forest (with 100 trees), a gradient boosting classifier (with 100 estimators), an SVM (with a radial basis kernel), and a four-layer DNN (trained using Adam for 100 epochs with a batch size of 100). First, we look at binary classification of CT vs AD. We then further split the AD group into two categories, Mild and Moderate, given MMSE scores above and below 10, respectively. Results for multi-class and binary classification are presented in Tables \ref{multiclassification} and \ref{binaryclassification} below. We report the F1 averages. The macro average assigns equal weight to each class, whereas the micro average accounts for the frequency of each class.

\begin{table}[h]
\caption{Moderate vs Mild vs CT. The three highest F1 macro scores are shown in bold.}
\centering
\begin{tabular}{|p{2cm}||p{1cm}p{1cm}|p{1cm}p{1cm}|p{1cm}p{1cm}|p{1cm}p{1cm}|}
\hline
                         & \multicolumn{2}{c|}{Random Forest} & \multicolumn{2}{c|}{Gradient Boosting} & \multicolumn{2}{c|}{SVM} & \multicolumn{2}{c|}{DNN} \\ \hline
                         & F1 \newline (macro)       & F1 \newline (micro)      & F1 \newline (macro)         & F1 \newline (micro)        & F1 \newline (macro)  & F1 \newline (micro) & F1 \newline (macro)  & F1 \newline (micro) \\ \hline
DB only                  & 			     66.53   & 69.73 & 66.47 & 68.81 & 67.50  & 71.56 & 20.18 & 43.40 \\ \hline
DB + WLS                & 			     69.08   & 92.12 & 68.74 & 91.33 & 66.23  & 88.98 & 55.81 &  86.65 \\ \hline
DB + WLS \newline (oversampled)    &   \textbf{70.26} & 90.05 & 68.46 & 90.31 & 68.66 & 89.53 & 69.45 & 90.84\\ \hline
DB + T2M             & 			      67.65  & 82.02 & 68.49 & 82.89 & \textbf{70.23}  & 81.58 & 62.70 &  79.82 \\ \hline
DB + T2M \newline (oversampled) & 69.17  & 76.89 & \textbf{70.14} &  81.33 & 57.87 & 72.97 & 58.46 & 64.91\\ \hline
\end{tabular}
\label{multiclassification}
\end{table}

\begin{table}[h]
\caption{AD vs CT. The three highest F1 macro scores are shown in bold.}
\centering
\begin{tabular}{|p{2cm}||p{1cm}p{1cm}|p{1cm}p{1cm}|p{1cm}p{1cm}|p{1cm}p{1cm}|}
\hline
                         & \multicolumn{2}{c|}{Random Forest} & \multicolumn{2}{c|}{Gradient Boosting} & \multicolumn{2}{c|}{SVM} & \multicolumn{2}{c|}{DNN} \\ \hline
                         & F1 \newline (macro)       & F1 \newline (micro)      & F1 \newline (macro)         & F1 \newline (micro)        & F1 \newline (macro)  & F1 \newline (micro) & F1 \newline (macro)  & F1 \newline (micro) \\ \hline 
DB only                  &                               79.49   & 79.63 & 76.50 & 75.93 & 80.00 & 80.56 & 36.91 & 58.49 \\ \hline
DB + WLS               &                             91.01   &  95.30 & \textbf{93.03} & 96.34 & 88.42 & 95.30 & 78.63 &  89.53  \\ \hline
DB + WLS \newline (oversampled)    &    \textbf{91.68} & 95.56 & 90.60 & 95.04 & 89.18 & 93.47 & \textbf{92.78} & 95.30  \\ \hline
DB + T2M           &                            84.97 &  87.23 & 86.47  & 88.55 & 82.89 & 87.22 & 79.10 & 72.92 \\ \hline
DB + T2M \newline (oversampled) & 82.42 &  80.70 & 83.60 & 85.09 & 78.99 & 80.26 & 77.19  & 73.25 \\ \hline
\end{tabular}
\label{binaryclassification}
\end{table}

\section{Discussion}

Effectively monitoring and assessing the linguistic symptoms of dementia automatically will have major potential impacts on health care. Among these is the ability to remotely assess cognitive function in mobility-reduced (and rural) individuals, which would considerably lessen the burden on healthcare workers. Clearly, using normative data greatly improves classification accuracy, and these improvements are generally maintained through class-balancing with ADASYN, although a Kruskal-Wallis test does not find statistical significance ($\chi^2(1)=2.34, p=0.13$). However, an $n$-way ANOVA reveals significant main effects of model ($F_3 = 5.83, p<0.01$), and task (binary v trinary, $F_1 = 8.91, p<0.01$), as well as interaction effects between the model and task ($F_3=3.43,p<0.05$) and between database and task ($F_2=6.79, p<0.01$), with database and oversampling as covariates. While oversampling does not typically {\em improve} estimates, it is important for verification, due to the massive class imbalance otherwise.

Considering the binary task in Table \ref{binaryclassification}, it is clear that adding WLS, rather than T2M, improves performance the most. There are two possible explanations for this; first, the demographics of WLS more closely resemble those of DB than T2M (especially in terms of age) and, secondly, T2M also includes some picture descriptions of images other than the Cookie Theft. Indeed, the features selected (as shown in Table \ref{features}) indicate what sets these normative data sets apart. While both data sets reveal similar differences to speakers with AD in the DB data set, in terms of grammatical features and semantic similarity, the latter is amplified in WLS, and T2M reveals more subjective or psycholinguistic differences. Interestingly, many lexical features (indicative of previous work that only used the DB data set \cite{Fraser2015}) ceased to be important when including the normative data. Future work will reveal if differences in grammatical construction may be indicative of slight cultural differences.

Because of the relatively small size of the DB dataset, we resorted to extensive feature engineering, and extracted a total of 567 features. The small amount of data also limited the performance of the DNN, since it only achieved comparable results when supplemented with a normative dataset. 

As expected, classification is easier in the binary case than the trinary case, especially due to relatively minor differences between speakers with mild versus moderate cognitive impairment. The very high micro F1-scores on the augmented datasets can be misleading without context, due to the high class imbalance. Selecting appropriate evaluation metrics, especially in this context, is paramount.

Ongoing work is focused on augmenting the T2M dataset, by collecting data from individuals with cognitive decline, and by also introducing a telephone-based interface. We are currently applying transfer- and multiview-learning on these and related data sets of pathological speech. 

\subsubsection*{Acknowledgments}

The Wisconsin Longitudinal Study is sponsored by the National Institute on Aging (grant numbers R01AG009775, R01AG033285, and R01AG041868), and was conducted by the University of Wisconsin.

\small
\bibliographystyle{plain}

\newpage

\section*{A. Features}

The top features derived in \S \ref{experiments} are presented in the tables below.

For the top syntactic features, the T-units are extracted using the Lu Syntactic Complexity analyzer \cite{Lu2010}, and the grammatical constituents are extracted from parse trees generated by the Stanford Parser \cite{Manning2014}. The semantic similarity measures consist of various metrics related to the cosine measure of similarity taken on each pair of utterances. For the top subjective measures, we consider reading norms, psycholinguistic features, and sentiment analysis. Reading norms were derived from the Flesch reading-ease score and Flesch-Kincaid grade level formula \cite{Kincaid1975}. Psycholinguistic measures were derived from LIWC \cite{Pennebaker2015} and Receptiviti \footnote{https://www.receptiviti.ai/liwc-api-get-started}. Measures for negative polarity are extracted from the Stanford Sentiment Analyzer \cite{Manning2014}.

\begin{table}[h]
\caption{Lexical features}
\begin{tabular}{|p{3.5cm}|p{9.5cm}|}
\hline
\textbf{Feature} & \textbf{Description} \\ \hline
Mean syl/word & The mean \# of syllables per word.\\ \hline
Mean word length & The mean \# of characters per word. \\ \hline
Adverbs & The total number of adverbs divided by the total number of words.\\ \hline
PRP ratio & The pronoun ratio, defined as the total number of pronouns divided by the total number pronouns and nouns.\\ \hline
\end{tabular}
\end{table}

\begin{table}[h]
\caption{Syntactic features}
\begin{tabular}{|p{3.5cm}|p{9.5cm}|}
\hline
\textbf{Feature} & \textbf{Description} \\ \hline
T & Total \# of T-units in the transcript, normalized by total \# of words. T-units are main clauses plus their dependent clauses.\\ \hline
S & Total \# of sentences. \\ \hline
$ROOT \ \rightarrow \ S$ & \# of parse tree root $\rightarrow$ declarative clause occurrences, normalized by the total \# of grammatical constituents, e.g., ``(ROOT (She is looking at the cookie jar))''.\\ \hline
$NP \ \rightarrow \ PRP$ & \# of noun phrases $\rightarrow$ personal pronoun  occurrences, normalized by the total \# of grammatical constituents, e.g., ``(NP (PRP he)) is looking at the cookie jar''.\\ \hline
$ADVP \ \rightarrow \ RB$ &  \# of adverbial phrases $\rightarrow$ adverb occurrences, normalized by the total \# of grammatical constituents, e.g. ``the water is (ADVP (RB still)) flowing''.\\ \hline
$S \ \rightarrow \ ADVP \ NP \ VP$ &  \# of declarative clause $\rightarrow$ adverbial phrase, noun phrase, and verb phrase  occurrences, normalized by the total \# of grammatical constituents, e.g. ``(S (ADVP evidently) (NP  they) (VP had been driving))''. \\ \hline
$S \ \rightarrow \ CC \ NP \ VP$ &  \# of declarative clause $\rightarrow$  coordinating conjunction, noun phrase, and verb phrase occurrences, normalized by the total \# of grammatical constituents, e.g. ``(S (CC And) (NP the curtains) (VP  are very distinct)).''\\ \hline
$S \ \rightarrow \ NP \ VP$ &  \# of declarative clause$\rightarrow$ noun phrase and verb phrase  occurrences, normalized by the total \# of grammatical constituents, e.g. ``(S (NP the sink) (VP is overflowing))''. \\ \hline
$VP \ \rightarrow \ VBZ \ VP$ &  \# of verb phrase $\rightarrow$ third person singular verb and verb phrase  occurrences, normalized by the total \# of grammatical constituents ``the mother (VP (VBZ is) (VP washing the dishes))''. \\ \hline

\end{tabular}
\end{table}

\begin{table}[h]
\caption{Semantic similarity measures}
\begin{tabular}{|p{3.5cm}|p{9.5cm}|}
\hline
\textbf{Feature} & \textbf{Description} \\ \hline
Mean cosine distance & Average cosine distance between each pair of utterances in the transcript, normalized by total number of unique utterance pairwise comparisons.\\ \hline
Min. cosine distance & Minimum cosine distance between each pair of utterances in the transcript. \\ \hline
Cosine cutoff (0.5) & Number of pairs of utterances whose cosine distance is less than 0.5, normalized by total number of unique utterance pairwise comparisons.\\ \hline
\end{tabular}
\end{table}

\begin{table}[h]
\caption{Subjective measures}
\begin{tabular}{|p{3.5cm}|p{9.5cm}|}
\hline
\textbf{Feature} & \textbf{Description} \\ \hline
Flesch & The Flesch reading-ease score evaluates the reading ease of a text based on the following formula: \newline
$F = 206.835 - 1.015 \frac{total \ words}{total \ sentences} - 84.6 \frac{total \ syllables}{total \ words}$ \newline \\ \hline
Flesch-Kincaid & The Flesch-Kincaid grade level formula evaluates the reading ease and presents a score corresponding to a U.S. grade level. The reading ease is evaluated based on the following formula: \newline
$FK = 0.39 \frac{total \ words}{total \ sentences} + 11.8 \frac{total \ syllables}{total \ words} +15.59$ \newline\\ \hline
LIWC $Apostro$ & The total \# of apostrophes, as identified by LIWC.\\ \hline
LIWC $ingest$ & The total \# of words that relate to the biological process of ingesting (eating), as identified by LIWC.\\ \hline
Receptiviti $sociable$ & Percentage score on how sociable the text seems, based on the Receptiviti system.\\ \hline
Receptiviti $family\_oriented$ & Percentage score on family oriented the text seems, based on the Receptiviti system.\\ \hline
Mean Stanford sentiment (negative) & The mean score of negative polarity for each utterance in the transcript.\\ \hline
\end{tabular}
\end{table}

\newpage
\section*{B. Model details}

Hyperparameters were tuned using grid search with 10-fold cross validation on the training set.

The random forest classifier fits 100 decision trees and considers $\sqrt{\# \ features}$ when looking for the best split. Each decision tree in the ensemble is built from a sample drawn with replacement

The gradient boosting classifier trains 1000 decision trees of depth 5, and a learning rate of 0.1. 

The SVM is trained with a radial basis function kernel ($\gamma = 0.001$).

The DNN used consists of four layers of 512 units. The $tanh$ activation function is used at each hidden layer, and a dropout of 0.1 is applied after each hidden each. The DNN is trained using Adam for 100 epochs and with a batch size of 100.  A learning rate of 0.1, and the cross-entropy loss are used.

\newpage
\section*{C. AUC values}
The receiver operating characteristic curve plots the true positive rate vs the false positive rate for different decision thresholds. The area under the curve (AUC) for each experiment is displayed in the tables below.

\begin{table}[h]
\centering
\caption{AUC values for binary classification. AUC values for the $AD$ class are computed.}
\label{aucbinary}
\begin{tabular}{|l|l|l|l|l|}
\hline
                        & Random Forest & Gradient Boosting & SVM & DNN  \\ \hline
DB only                 & 0.89          & 0.89              &   0.89 & 0.76 \\ \hline
DB + WLS                & 0.99          & 0.99              &  0.97   & 0.98 \\ \hline
DB +  WLS (oversampled) & 0.99          & 0.99              &  0.83   & 0.98 \\ \hline
DB +  T2M               & 0.95          & 0.97              & 0.93    & 0.83 \\ \hline
DB + T2M (oversampled)  & 0.89          & 0.96              & 0.68    & 0.86 \\ \hline
\end{tabular}
\end{table}

\begin{table}[h]
\centering
\caption{AUC values for the multi-class problem. AUC values are computed in a one-vs-all approach. Values for the undersampled classes $Moderate$ and $Mild$ are reported.}
\label{aucmulti}
\begin{tabular}{|p{2cm}||p{1cm}p{1cm}|p{1cm}p{1cm}|p{1cm}p{1cm}|p{1cm}p{1cm}|}
\hline
                        & \multicolumn{2}{l|}{Random Forest} & \multicolumn{2}{l|}{Gradient Boosting} & \multicolumn{2}{l|}{SVM} & \multicolumn{2}{l|}{DNN} \\ \hline
                        & Moderate           & Mild          & Moderate             & Mild            & Moderate      & Mild     & Moderate      & Mild     \\ \hline
DB only                 & 0.87               & 0.70          & 0.84                 & 0.69            &               0.91 & 0.71         & 0.80          & 0.60     \\ \hline
DB + WLS                & 0.96               & 0.96          & 0.96                 & 0.96            &  0.97             & 0.96         & 0.93          & 0.95     \\ \hline
DB +  WLS \newline (oversampled) & 0.95               & 0.95          & 0.97                 & 0.96            &     0.76          & 0.84         & 0.92          & 0.95     \\ \hline
DB +  T2M               & 0.96               & 0.86          & 0.96                 & 0.85            &    0.92           &      0.89    & 0.80          & 0.78     \\ \hline
DB + T2M \newline (oversampled)  & 0.94               & 0.83          & 0.94                 & 0.84            &   0.85            & 0.75         & 0.95          & 0.76     \\ \hline
\end{tabular}
\end{table}

\end{document}